# Computational conceptual history of scientific concepts

## From early digital methods to LLMs

*Michael Zichert and Arno Simons*

## 1. Introduction

This article zooms out to ask what it means to use large language models (LLMs) for computational conceptual history of scientific concepts. Our aim is to situate LLMs within the longer history of digital approaches to concept analysis in the history, philosophy, and sociology of science (HPSS), examine what these models add to existing methods, and review recent case studies that already employ them. This perspective allows us to ask not only what new possibilities LLMs open, but also how they inherit longstanding problems from both conceptual history and earlier attempts to study concepts computationally. In this article, we show how LLMs both extend and reshape computational conceptual history by revisiting its core methodological issues: corpus construction, concept operationalisation, model and training data choice, and the linked challenges of evaluation and interpretation.

The analysis of scientific concepts has long been a core concern in HPSS, where concepts are treated as historically layered, socially situated, and epistemically significant components of scientific practice rather than as fixed semantic units. Traditions of conceptual history and historical epistemology, as well as later developments in the discursive, practical, and material turns, have shown that concepts gain meaning through their use in specific research settings, institutional contexts, and socio-technical arrangements. Scholars such as Fleck (1979[1935]), Kuhn (1962), Rheinberger (1997), Hacking (1999), Pickering (1999), and Daston and Galison (2007) have emphasised that conceptual change often tracks deeper shifts in scientific problems, practices, and forms of reasoning, and that key terms can carry multiple, sometimes conflicting meanings that shape how scientists interpret phenomena.

Recent work has renewed interest in conceptual history within the field (Schauz, 2015; Müller and Schmieder, 2016; Schauz, 2025), alongside explicit calls to connect conceptual analysis more systematically to quantitative and computational methods in conceptual history more broadly (Wevers and Koolen, 2020; Marjanen, 2023). In parallel, historians of science have increasingly pointed to the methodological potential of large-scale textual and bibliometric evidence and have called for computationally grounded histories of science (Laubichler, 2019; Lalli, 2023). Most recently, LLMs have intensified these discussions by making new forms and scales of contextual modelling available, prompting HPSS scholars to explore how these models can be integrated into concept historical research designs, as reflected by this volume and the contributions in this part of the book. For a more comprehensive survey of LLMs for HPSS, see our earlier work (Simons et al., 2026).

The article has two main parts. Chapter 2 reconstructs computational conceptual history before LLMs by bringing together the three strands of work: early digital methods in HPSS, distributional approaches from digital history and related research, and lexical semantic change detection. It then sums up the main challenges and opportunities, focusing on corpus construction, operationalization and modelling choices, and evaluation and interpretation. Chapter 3 turns to the era of LLMs. It starts with a short introduction to LLMs, reviews LLM-based work on lexical semantic change detection, and discusses relevant case studies in HPSS that make use of LLMs. It then revisits the earlier methodological questions, showing how issues like corpus construction, model choice and training data, operationalization trade-offs, and evaluation and interpretation play out in LLM-based workflows.

## 2. Computational conceptual history before LLMs

### 2.1 Early digital approaches to modeling conceptual change

#### 2.1.1 Early digital methods in HPSS

Alongside the well-established qualitative tradition of conceptual history in HPSS, scholars, especially in the scientometrics community, have long explored digital methods for conceptual history and analysis. One of the most influential early contributions is Small's work on co-citation analysis (1973). Co-citation refers to the frequency with which two publications are cited together by later authors. Small links this technique to a broader claim about meaning in scientific literature. He argues that frequently cited papers can function as concept symbols (1978), serving as shorthand for ideas, methods, or findings that a research community recognizes as central. From this perspective, co-citation is useful because it traces how these symbolically important contributions are connected in practice and therefore reflects the conceptual associations that the wider population of researchers sees between key works. As these perceptions and citing habits shift, co-citation patterns change as well, offering a dynamic view of how a field's core ideas are linked and reorganized over time. If highly cited papers are taken to mark major concepts or methods, then co-citation analysis can map the evolving structure of these central elements with considerable detail (Small, 1999; Boyack et al., 2005; Boyack and Klavans, 2014).

Building on this foundation, Callon et al. (1983, 1986) introduce co-word analysis as a complementary computational strategy for mapping conceptual structures through language use in the scientific literature. Rather than inferring conceptual relations from shared references, this approach traces the co-occurrence of keywords and index terms to reveal how research communities connect problems, methods, and objects of inquiry within evolving discourses. As a result, co-word maps can highlight emerging themes, shifting alliances between topics, and the consolidation of new research areas, offering a vocabulary-centered view of conceptual change over time (Rip and Courtial, 1984; Ding et al., 2001).

A third pre-LLM computational approach to conceptual history in HPSS builds on topic modeling, in particular Latent Dirichlet Allocation (LDA) and its extensions (Blei et al., 2003; Blei and Lafferty, 2006). Topic models represent documents as mixtures of latent topics and topics as probability distributions over words, so that each topic can be read as a recurring constellation of terms that approximates a conceptual or thematic structure. By fitting such models to large scientific corpora and tracking how topics change in prevalence and internal composition over time, researchers have used LDA-based methods to investigate the emergence, transformation, and decline of research areas, thus providing a scalable way to study conceptual dynamics in science (Griffiths and Steyvers, 2004; Blei and Lafferty, 2007; Hall et al., 2008).

For critical comparisons between these different methods, see for example Boyack and Klavans (2010), Leydesdorff and Nerghes (2017), Benz et al. (2025), and Xie and Waltman (2025).

### 2.1.2 Digital Begriffsgeschichte

In parallel to these HPSS approaches, digital humanities and digital history have developed computational conceptual history in dialogue with Begriffsgeschichte (Koselleck, 2002; Koselleck and Richter, 2011), drawing on co-occurrence analysis and, later, distributional models to trace changing patterns of use. In Begriffsgeschichte, concepts are treated as historically layered semantic structures whose meanings accumulate, fracture, and compete over time, and which can be reconstructed to trace broader social and intellectual transformations. This perspective shares with conceptual history in HPSS the commitment to historical situatedness and contextual interpretation. Where Begriffsgeschichte tends to be more explicit, however, is in treating polysemy as a central methodological issue and in emphasizing the distinction between the words that appear in texts and the concepts behind them. This also brings with it a comparatively developed account of what concepts are and why the term–concept relation is methodologically fraught. By contrast, much pre-LLM computational work in HPSS has operationalized concepts through comparatively stable proxies (highly cited papers, index terms, topic labels) and mapped the resulting structures onto concepts in a more pragmatic, case-specific way, often without foregrounding the term–concept relation at the same level of theoretical depth. As a result, polysemy and sense competition tend to remain implicit or get smoothed over rather than becoming a first-order methodological concern.

Over the past decade, scholars in this field have increasingly tried to operationalise conceptual history with pre-LLM digital methods, especially through co-occurrence and early word embedding techniques to track shifts in contextual patterns and semantic neighborhoods across large corpora. Biemann and Friedrich (2016) highlight both the promise of large digitised corpora for mapping contexts of use and the difficulty of semantic ambiguity when meaning is inferred from distributional patterns. Schwandt (2018) similarly uses tools such as Voyant (Sinclair and Rockwell, 2016) to unsettle mod-ern readings of historical concepts, while stressing how hard it is to compare uses across periods and genres. Gavin et al. (2019) explore a hybrid approach using vector semantics in computer-assisted close reading. Working with Word2Vec embeddings (Mikolov et al., 2013), they treat distributional representations not as stand-alone models of meaning but as tools that can support interpretation. In this way, embeddings serve to draw attention to certain patterns in the corpus rather than to replace close reading.

Where Gavin et al. highlight the interpretive potential of embeddings, Sommerauer and Fokkens (2019) focus on their methodological risks. Using "racism" as a case study, they show how distributional semantic models trained on the same corpus can yield divergent results depending on architecture, parameters, and corpus composition, demonstrating the need for controlled experimental design if such methods are to support historical claims. The most systematic pre-LLM account is Wevers and Koolen (2020), who use static embedding models such as Word2Vec (Mikolov et al., 2013) and GloVe (Pennington et al., 2014) on Dutch newspaper corpora to highlight opportunities to trace and measure semantic and conceptual change via shifts in vector positions and neighbourhoods. At the same time, they emphasise that embedding based results require interpretive grounding and validation, since the operationalization of concepts through lexical proxies, the size, quality, and bias of historical corpora, and the difficulties of evaluating model quality all require careful methodological reflection.

### 2.1.3 Pre-LLM approaches to lexical semantic change detection

Another tradition of studying shifting word and concept meanings with digital methods arises from research on lexical semantic change detection (LSCD). Emerging within computational linguistics, computer science, and natural language processing, LSCD uses many of the same distributional and embedding-based techniques as digital conceptual history, but typically focuses on technical implementation, the construction and evaluation of models and datasets, and systematic benchmarking rather than on substantive historical interpretation. Methodologically, its goal is to formalize and measure lexical semantic change in ways that allow different models and approaches to be compared. In recent years, however, LSCD techniques have increasingly been taken up and adapted within digital conceptual history, where they are combined with closer attention to historical context, sources, and interpretation.

Early LSCD approaches relied on methods such as dynamic topic modelling (Blei and Lafferty, 2006), statistical co-occurrence analysis (Gulordava and Baroni, 2011), and graph-based techniques (Mitra et al., 2014) to capture changes in the contexts in which words appear. With the rise of distributional semantics, vector-based methods became dominant. A central distinction is between static word embeddings, which represent each word type in a corpus as a single fixed vector (e.g., Word2Vec, GloVe), and more recent contextualized approaches, in which every word occurrence is represented by its own unique vector through the use of LLMs like BERT (Devlin et al., 2018) or GPT (Radford et al., 2018). Static embeddings make it comparatively straightforward to measure shifts in the predominant use of a word by comparing vectors across time slices, but they struggle with highly polysemous items, where multiple senses are collapsed into a single representation.

Within the pre-LLM, static embedding approach, Hamilton et al. (2016) propose influential methods for training separate Word2Vec models on different time slices, aligning the resulting spaces, and quantifying change by measuring cosine distance between a word's vectors across periods, which they use to suggest broader regularities linking semantic change to frequency and polysemy. Dubossarsky et al. (2017) subsequently caution that some such patterns may reflect artefacts of the models and evaluation procedures rather than genuine linguistic regularities, highlighting inherent biases in word representation models and the need for careful baselines and controls. Kutuzov et al. (2018) provide a comprehensive survey of pre-LLM SCD work, summarizing tasks, datasets, and methodological choices. Taken together, this tradition has strongly shaped the toolkits that conceptual historians now draw on, but as HPSS scholars we need to understand how these methods work, how to adapt them to our own questions and corpora, and how to combine them with the contextual interpretation and conceptual analysis at the core of our field.

## 2.2 Opportunities and challenges of pre-LLM computational conceptual history

Building on the three strands introduced in 2.1, this section draws together their key lessons to identify the main opportunities and recurring challenges of pre-LLM computational conceptual history, from the promise of scale to the constraints of corpus construction, concept operationalization, evaluation, and historically accountable interpretation.

### 2.2.1 Text as data and corpus construction

The availability of large digitized and, increasingly, born-digital text collections has often been framed as the decisive opportunity for computational approaches to conceptual history, because it allows scholars to study conceptual dynamics at scales of time and corpus size that were not feasible before (Biemann and Friedrich, 2016; Wevers and Koolen, 2020). Digitization makes sources easier to access and aggregate, but it also encourages historians to encounter their material through search, for example via keyword queries, indices, and frequency tools, so that individual word forms become common entry points into archives that were formerly approached through documents, genres, or debates. This invites new kinds of questions about how the relationship between words and concepts can be quantitatively operationalized and how linguistic change can be measured across large text corpora. As Marjanen (2023) notes in a wider discussion of quantitative approaches to conceptual history, this development is broadly good news for the field, but it also requires researchers to reflect carefully on what quantification captures and how large-scale linguistic patterns should be interpreted.

At the same time, what can be claimed from such large-scale analyses is constrained by decisions about corpus construction. What gets digitized, what is missing, how texts are segmented, and what metadata is available for filtering and comparison all shape what can reasonably be concluded from the analysis. Digitization is uneven across regions, institutions, and genres, and corpora are also uneven over time: recent decades often dominate, while earlier periods can be sparse. In addition, OCR quality, historical spelling variation, and corpus bias are primary methodological issues for embedding based historical analysis (Wevers and Koolen, 2020). Pre-LLM distributional approaches are also sensitive to preprocessing decisions, such as tokenization, lemmatization and stopword removal. HPSS corpora amplify these issues because scientific language is often highly domain-specific and heterogeneous (e.g., dense technical terms, abbreviations, and shifting terminological conventions). In addition, conceptual work is often distributed across different types of sources (e.g., articles, textbooks, grant proposals, review papers, lab manuals, etc.), and can also be encoded in non-textual elements (e.g., formulae, tables, figures, code, and diagrams) that are typically stripped during digitization and preprocessing when employing pre-LLM methods, further increasing the problem of what purely text-based models can measure.

### 2.2.2 Operationalization and modeling choices

One central methodological concern for computational conceptual history is the operationalization, or modeling, of concepts. To employ digital tools, researchers must translate historically layered and contested semantic structures into measurable proxies. This translation is never neutral and is further complicated by the fundamental word concept gap emphasized in conceptual history (de Bolla et al., 2019; Marjanen, 2023). As discussed in 2.1, pre-LLM approaches operate on words, citations, keywords, or topic distributions, and they infer conceptual meaning from these different surface forms. In practice, this often means relying on lexical proxies, for example seed terms, keyword sets, anchor papers, or topic word distributions, to stand in for broader conceptual structures. This can be productive, but it also makes results sensitive to the initial proxy choice and to whether the proxy actually captures the relevant uses, genres, and debates in the corpus.

In pre-LLM conceptual history, three recurring strands of operationalization can be observed, each with specific modeling trade-offs. Scientometric networks are closely tied to scholarly behavior and yield useful representations, but they primarily track patterns of association and community organization rather than conceptual content. Topic models enable scalable thematic mapping that is useful for exploration and periodization, but whether topics correspond to concepts remains contested (Chang et al., 2009). Early embedding methods align most closely with intuitions of conceptual meaning based on context and semantic neighborhoods, but they introduce difficulties with polysemy and aggregation. Pre-LLM approaches rely on static word embeddings that summarize all occurrences of a word into a single representation, which can flatten disputes and sense variations that are often central to conceptual historical analysis. A further limitation specific to HPSS is that scientific concepts are often considered material-semiotic: they stabilize not only through words, but also through instruments, experimental systems, and standardizing infrastructures, so conceptual change is not always legible as lexical change alone (Müller and Schmieder, 2018).

### 2.2.3 Evaluation and interpretation

Evaluation is a persistent challenge in pre-LLM computational conceptual history because most approaches lack straightforward ground truths and because in most humanities-aligned areas there are no widely-accepted evaluation datasets, an issue that is especially acute in HPSS (see the introduction to this volume, Simons et al., 2026). Topic models, for example, can yield substantially different topic structures depending on hy-perparameters and preprocessing, raising the question of whether an inferred topic represents a stable conceptual formation or an artefact of the modeling setup. Static embedding workflows also depend heavily on corpus composition and parameter settings, so robust evaluation typically requires careful baselines, controlled comparisons, and triangulation of results (Dubossarsky et al., 2017; Sommerauer and Fokkens, 2019).

Across all these approaches, careful historical interpretation remains essential. In computational conceptual history, outputs such as frequency curves, co-citation clusters, co-word networks, or embedding shifts are rarely historical claims on their own. Instead, they point to patterns that have to be turned into arguments. Interpretation is this conversion step: deciding what a pattern can count as evidence for, which alternative explanations need to be checked and ruled out, and how the result connects to actors, instruments, practices, or institutions. LSCD often concentrates on measuring change and benchmarking models and typically leaves these interpretive steps outside its scope. For humanities and HPSS scholars, however, interpretation is the core of conceptual analysis, and pre-LLM work therefore kept interpretive responsibility firmly with the researcher, whether through a hybrid approach as in Gavin et al. (2019) or through embedding-based conceptual history that explicitly insists on interpretive validation as in Wevers and Koolen (2020).

## 3. LLMs for computational conceptual history

Building on the challenges and opportunities discussed in the previous chapter, the next two sections outline what LLMs can contribute to computational conceptual history, which problems they inherit, and which gaps they may help to close.

### 3.1 Recent advances and and HPSS case studies

#### 3.1.1 Short primer on LLMs

As outlined in the introduction to this volume (Simons et al., 2026) and in our survey (Simons et al., 2026), we use LLMs to refer to neural language models built on the transformer architecture (Vaswani et al., 2017). Transformers build on and extend components introduced in earlier neural network models, most importantly attention mechanisms, encoder and decoder modules, and contextualized word representations, and integrate them into an architecture that scales effectively to very large datasets and parameter counts and produces rich representations of language.

Within this family, it is useful to distinguish two dominant architectural and training paradigms (see the Introduction to this volume for more details, Simons et al., 2026): Decoder-based, or generative models, often associated with chat-based systems such as ChatGPT, are trained with a next-token prediction objective, where given a text input sequence, the model learns to predict the most likely continuation (Radford et al., 2018). Encoder-based models, by contrast, are typically trained with masked language objectives, where the model learns to reconstruct tokens that have been hidden from an input sequence. The paradigmatic example here is BERT (Devlin et al., 2018), which is why they are also being referred to as BERT-like models. Because the encoder is oriented bidirec-tionally, it can use information from both the left and right context when representing a token, which makes it especially strong at modelling local meaning and sentence-level structure. Both model types produce context-sensitive vector representations for each token occurrence, so-called *contextualized* word embeddings (CWEs). Unlike *static* word embeddings, which assign a single vector to a word type across a whole corpus, CWEs are conditioned on the surrounding sentence or passage, so the same word form can have different representations in different contexts.

The distinction between decoder-based and encoder-based models matters for computational conceptual history because it tends to support different applications. Generative models are useful for text production and can, through prompt engineering, support structured extraction and forms of interpretive assistance. Encoder-based models, meanwhile, remain central for measurement oriented workflows, where CWEs serve as features for tasks such as lexical semantic change modelling, classification, or sequence labelling. One key advantage here is they can be efficiently adapted through task-specific fine tuning or domain adaptation on comparatively small datasets tailored to the historical or disciplinary question at hand. Generative models, by contrast, are often adapted at use time through in context learning, for example via few-shot or multi-shot prompting, or through retrieval-augmented generation (RAG), where relevant passages are retrieved from an external collection and included in the prompt to ground and influence the model's output (Gao et al., 2024).

### 3.1.2 Lexical semantic change detection (LSCD) using LLMs

LSCD with transformer models was initially driven mainly by encoder-based models, since their bidirectional training objective makes them well suited for detecting meaning shifts. A common workflow begins by extracting all occurrences of a target word from a diachronic corpus and representing each occurrence with a CWE extracted from a BERT-like model that is often domain adapted or further pretrained on the relevant historical material. These CWEs are then aggregated in two main ways: form-based approaches pool all occurrences into a single time-specific representation, while sense-based approaches use clustering methods to group the vectors into sense-specific groups. These time-specific representations can then be used to assess the amount of lexical change through distance-based measures such as cosine distance, or information-theoretic indicators such as Shannon entropy. This makes it possible to analyse different types of semantic change, most prominently shifts in a word's dominant meaning and changes in its degree of polysemy. Many recent studies both build on and further develop this pipeline (e.g., Martinc et al., 2020; Montariol et al., 2021; Kutuzov et al., 2022; Cassotti et al., 2023), and the survey by Periti and Montanelli (2024) offers the most systematic review of approaches, benchmarks, and model choices.

More recently, decoder-based or generative models have entered LSCD research along two somewhat different lines. First, they can be used directly for change detection through prompting, for example by asking the model for sense judgements, semantic relatedness decisions, or lexical substitutes. Periti et al. (2025) test a range of prompting strategies, including zero-shot and multi-shot setups with ChatGPT, on both long-term and short-term benchmarks for semantic change. They conclude that this direction is promising, but that encoder-based approaches with BERT-like models still achieve stronger performance, especially for fine-grained, short-term distinctions. Second, generative models are increasingly being used in hybrid approaches. Rather than replacing encoder-based measurement, they are applied to generate historically plausible, sense-specific example sentences that serve as synthetic, annotated diachronic datasets for training and evaluation, directly targeting the problem of sparse historical data. In these approaches, semantic change is still assessed with an encoder-based model. Cassotti and Tahmasebi (2025a), for instance, show how a Llama-based generative model can be used for sense-specific historical usage generation at scale, and how such data can be used for the standard downstream LSCD workflow. In a second paper that focuses on LSCD for highly polysemous target words (Cassotti and Tahmasebi, 2025b), they extend this idea by using decoder-based models not only to generate usages, but also to produce context- and time-specific sense definitions for the induced sense clusters, which are then used as anchors for the subsequent analysis.

#### 3.1.3 HPSS case studies using LLMs

In recent years, HPSS scholars have increasingly adopted LLMs to computational analysis of scientific concepts. While these studies often draw on methods developed in LSCD, they also integrate other models of conceptual change and adapt them to the specific problems and questions raised by scientific concepts.

In this volume, several chapters show how HPSS scholars are adapting LLM-based methods to the analysis of scientific concepts. Malaterre and Lareau (2026) discuss how both encoder-based and generative models can be used to analyse the ways scientists deploy epistemic concepts in the biomedical domain. Focusing on epistemic markers such as "theory", "model", and "explanation", they propose a hybrid workflow in which LLMs are used to expand and refine lexicons of epistemic markers, classify statements in context, and map discipline-specific epistemic framings at scale. Ahmadi (2026) uses CWEs from BERT to compare how consistently terms are used in sociology versus astrophysics journal articles. She computes a field-level Semantic Uniformity Score by averaging cosine similarity across each term's own contextualized occurrences and aggregating across terms, interpreting higher similarity as more stable, standardized usage and lower similarity as more context-dependent use shaped by polysemy, topic, syntax, or genre. She finds astrophysics to be more semantically uniform than sociology. Aguilar Valdez et al. (2026) sketch how the generation of a graph structure of concepts, which could then be enriched with an LLM-based analysis, could take conceptual analysis to the next level, while also using their study to familiarise readers with key concepts for computational approaches to semantic change. Simons (2026) uses the term "Planck" as a deliberately polysemous test case with well-known senses for computational conceptual analysis in astrophysics, combining several complementary methods. Using a domain-adapted BERT model that is further pretrained on a large astrophysics and high energy physics corpus (Simons, 2024), he evaluates how well contextualized embeddings separate these senses through supervised sense prediction with sense prototypes, as well as through unsupervised clustering and cluster quality measures. Building on the resulting sense structure, he then traces lexical semantic change from 1990 to 2022 by tracking how the distribution of senses shifts over time and by combining sense based- and form-based indicators.

In an earlier study, Kleymann et al. (2022) approach the use of the concept of theory in digital humanities journals by combining a semasiological perspective with an onomasiological one. In their semasiological case study, they build dictionaries of theory-related references and use frequency based and co-occurrence analyses to map how specific theoretical frameworks are used across the corpus. In their onomasiological case study, they follow the LSCD pipeline, fine-tuning a BERT model on the journal corpus, extracting CWEs for all occurrences of "theory" and related epistemic terms such as "model" and "method", and then comparing period specific and aggregated representations via cosine similarity to trace how the concept's semantic neighbourhood, and with it its use and meaning, have changed over time.

Lastly, Zichert et al. (2025) trace the conceptual history of the virtual particle concept by using "virtual" as a linguistic marker and analysing its usage across a large physics corpus spanning 1924 to 2022. They domain adapt a pretrained BERT model on this corpus, extract CWEs for each occurrence of "virtual", and apply a range of form-based and sense-based LSCD methods to track both shifts in dominant meaning and changes in the degree of polysemy over time. The approach is complemented by dependency parsing, which identifies the nouns most often used with "virtual", and these syntactic patterns are then used as an interpretive cross check on the LLM-based methods. Across all of these HPSS case studies, the dominant methodological approach remains the use of CWEs generated by encoder-based LLMs, while decoder-based approaches do not yet play a comparable role.

## 3.2 Opportunities and challenges for the LLM-era

Building on the developments and case studies surveyed in 3.1, this section draws their key lessons together to update the main opportunities and recurring challenges of computational conceptual history in the LLM era, comparing them to the pre-LLM landscape and highlighting how encoder-based and decoder-based models, and the model choices they entail, reshape questions of corpus and training data, concept operationalization and modelling trade-offs, and the tasks of evaluation and interpretation.

### 3.2.1 Corpus construction and datasets

Corpus construction remains central in the LLM era, but it has become more layered. Many pre-LLM issues, like uneven digitization, missing genres, or sparse early periods, remain and still shape what can be measured in HPSS conceptual history. What changes is that LLM-based workflows also inherit the biases and blind spots of the datasets the models are trained on. This is especially true for large-scale generative models where analysis is influenced not only by the researcher-curated corpus but also by the often opaque pre-training datasets. As a result, questions of infrastructure, access, and power become more relevant: who controls model training and updates, who can afford computation, and how environmental and financial costs shape what can or should be researched (Lang, 2026). For conceptual history and HPSS more broadly, this increases the need for open-science infrastructures and transparent alternatives (Valleriani, 2025).

At the same time, LLMs also offer significant opportunities regarding text corpora: They are far more robust to raw text than earlier pipelines, reducing the need for heavy preprocessing that can erase rhetorically and historically meaningful signals. More importantly for HPSS, domain adaptation has become a practical and standard task. This is especially true for encoder-based models, which can be effectively further pretrained or fine-tuned on discipline-specific corpora to better capture the specialized vocabulary and shifting terminological conventions of scientific language (e.g., Simons, 2024; Zichert et al., 2025). Another promising direction regarding corpora is temporal adaptation. Because most current LLMs are not trained with historical questions in mind, they risk projecting contemporary meanings into earlier texts. Time-sensitive fine-tuning and explicit temporal modeling aim to counteract this tendency (Underwood, 2025; Büttner, 2026; Meding and Daugs, 2026). The problem of historical sparse data can also be addressed through the use of generative LLMs that can help support corpus work itself: assisting annotation and coding, improving search within conceptual fields, as well as generating synthetic data to support training or evaluation in sparse domains (Cassotti and Tahmasebi, 2025a; Danilova et al., 2026).

### 3.2.2 Operationalization

Operationalization remains a central concern in LLM based computational conceptual history, but the range of modelling strategies and outputs has expanded. The most important shift is from static to contextualized word embeddings, which most current case studies generate through encoder-based models (e.g., Kleymann et al., 2022; Zichert et al., 2025; Simons, 2026). CWEs represent individual occurrences and therefore handle polysemy more directly, which makes it possible to model time specific uses through clustering, prototype-based sense prediction, or related word-sense disambiguation workflows, and then assess change either at the level of dominant senses or at the level of shifting sense distributions, which can be used to approximate changes in polysemy over time. These contextualized approaches also introduce new modelling trade-offs: results depend on choices such as which model layers to use, how token representations are merged, and how corpora are divided into time periods, while sense modeling requires decisions about clustering methods, the number of senses, and how clusters are labelled (Periti and Montanelli, 2024). Since many workflows still aggregate representations over time slices or clusters, CWEs improve access to polysemy but do not by themselves determine what a historically meaningful sense is.

Decoder-based models, on the other hand, modify operationalization in a different way. Mostly employed in hybrid workflows, they can be used to extract candidate definitions or typical usages of a term through retrieval-augmented generation (Gao et al., 2024), to build and refine coding schemes (Dunivin, 2025), or generate summaries of how a concept is discussed within a time text or subcorpus (Zhang et al., 2024). In LSCD, this direction appears both in prompt engineering based approaches that ask the model to judge whether a term is used with the same meaning across different contexts, or to propose substitute words for a given usage (Periti et al., 2025), and in synthetic data generation strategies that produce sense and time specific example usages of a concept or term (Cassotti and Tahmasebi, 2025a). These workflows also introduce new modelling choices that require explicit documentation, including model use, prompt design, in context examples, and output schemas. Much of this work is still in an early stage, but it is likely to become more important as generative models improve and as scholars develop more reliable hybrid pipelines for historical analysis.

Despite these advances, the long-standing operationalization problem in computational conceptual history remains relevant: concepts are bound to words, but they are not reducible to them, and linguistic proxies only capture part of what conceptual historians ultimately care about (Koselleck, 1985; Marjanen, 2023; de Bolla et al., 2019). In HPSS, this point is especially important, since scientific concepts are often understood as material-semiotic constructs that are jointly shaped by language, scientific practices, and material arrangements (Müller and Schmieder, 2018). Rheinberger (1997), for example, argues that concept histories of epistemic objects cannot be reconstructed from texts alone, since diagrams, formulae, and the material organisation of research practice are part of how scientific meaning is produced and transformed. This is where the rise of multimodal LLMs (Wu et al., 2023; Yin et al., 2024) marks a potential new direction. By integrating non-textual data such as visual and oral communication, they may enable future work to connect these dimensions more systematically, for example by tracing how textual uses of a concept co-evolve with its diagrammatic and mathematical representations. At the same time, this is still early work, that, besides multimodal LLMs, requires suitable multimodal corpora, improved digitisation of figures and formulae, and reliable alignment between textual and non-textual elements.

### 3.2.3 Evaluation and interpretation

Evaluation remains an important issue for computational conceptual history using LLMs. Most of the workflows discussed in 3.1 already include some form of evaluation, whether through established LSCD benchmarks or through the use of auxiliary quantitative methods such as frequency and co-occurrence measurements or dependency parsing that can support or challenge embedding-based findings. This need is amplified by the fact that LLM-based embeddings are more opaque, more “black boxed”, than most of the earlier quantitative methods in conceptual history, where assumptions and operationalization steps were often easier to inspect. As long as we have not established in detail how reliably these models capture conceptual dynamics, triangulation will remain the default evaluation strategy, and strong studies combine multiple quantitative lenses while pairing them with qualitative validation. As Marjanen (2023) argues, machine learning outputs only become meaningful when qualitative work clarifies what a measure is actually sensitive to (e.g., genre shifts, general linguistic trends, or changing publication volumes) and thereby helps distinguish conceptual dynamics from artefacts of the corpus or model (compare the discussion by Ahmadi, 2026, of whether high variability in embeddings signals semantic ambiguity, topical shifts, syntactic variation, or genre differences). This is especially important for HPSS, where we rarely have stable ground truths and where domain- and period-specific benchmark datasets are rare, and it is therefore unlikely that evaluation will center on a single standardized metric any time soon. For many (often commercial) generative models, this evaluation problem is sharpened by familiar reliability issues such as hallucinations, non-transparent model version updates and limited training-data transparency, that further complicate reproducibility (see Lang, 2026).

Interpretation remains the core operation of HPSS computational conceptual history, even in the LLM era. LLM-based workflows can support interpretive work in (at least) two complementary modes: first, they can test qualitative claims by empirically checking whether proposed shifts, stabilizations, or conceptual reorientations are visible across much larger corpora and longer timeframes than traditional case studies allow. Second, they can be used heuristically to discover interesting periods or candidate passages that call for closer analysis. As discussed earlier, LLMs in the form of generative models, can also support interpretation within hybrid workflows, for example by helping to build and refine coding schemes, extracting candidate definitions or typical uses of a term from retrieved contexts, and generating structured summaries of how a concept is framed within a time slice. At the same time, the increasing use of such systems raises a distinctive interpretive risk: even when models are used as assistants, their outputs can subtly shift interpretive control away from scholars (Khutsishvili, 2026). It remains to be seen how far such workflows can be pushed, since so far no case studies in computational conceptual history in HPSS have done so on a large scale.

## 4. Conclusion

In this contribution we have argued that LLM-based computational conceptual history in HPSS continues and reconfigures earlier digital approaches to modelling conceptual change, from co-citation methods to topic models and distributional semantics. LLMs expand what can be operationalised at scale by offering richer context-sensitive representations for tracing semantic and conceptual change through encoder-based models, and by enabling new hybrid workflows through the addition of generative models. At the same time, the core methodological questions remain, but LLMs shift how they have to be addressed, opening new possibilities while also adding new challenges. Corpus construction for analysis as well as model training, evaluation, concept operationalization, and domain as well as temporal adaptation are still decisive for what can be claimed. In practice, historically-accountable conceptual analysis continues to depend on triangulation, combining LLMs with established quantitative techniques and qualitative validation, with LLMs functioning as one component in a multi-method toolbox rather than as a self-sufficient solution.

Several avenues for future work in the field stand out. First, HPSS conceptual history needs more case studies that test generative models in hybrid workflows alongside encoder-based models, and that explore when, where, and for what tasks standalone generative approaches add value beyond supportive roles within conceptual history pipelines. Second, multimodal modeling opens a path toward operationalizations closer to how many HPSS scholars understand scientific concepts, as strongly connected with instruments, diagrams, formulae, and other material and representational practices, but this will require new corpora and reliable alignment between textual and non-textual elements. Third, progress across these lines of work depends on open and transparent infrastructures, open and HPSS-specific models and datasets, clear documentation stan-dards, and shared evaluation resources that make results reproducible and meaningfully comparable across studies and over time.[1]

1 This chapter was written with support from large language models (LLMs). All model-generated text was reviewed and, where necessary, rewritten by the authors, who remain fully responsible for the final version. For details on the use of LLMs in this volume, see the statement in the volume's introduction.